\documentclass[letterpaper, 10 pt, conference]{ieeeconf}

\IEEEoverridecommandlockouts                              

\usepackage{algorithm2e}
\usepackage{graphicx}
\usepackage{amsmath,amssymb}
\usepackage[dvipsnames]{xcolor}
\usepackage{url}

\title{\LARGE \bf
Using Contrastive Samples for Identifying and Leveraging Possible Causal Relationships in Reinforcement Learning
}

\author{Harshad Khadilkar$^{1,2}$ and Hardik Meisheri$^1$
\thanks{$^{1}$TCS Research
        {\tt\small harshad.khadilkar@tcs.com}}%
\thanks{$^{2}$IIT Bombay
        {\tt\small harshadk@iitb.ac.in}}%
}

\begin{document}
\maketitle
\thispagestyle{empty}
\pagestyle{empty}




\begin{abstract}
  A significant challenge in reinforcement learning is quantifying the complex relationship between actions and long-term rewards. The effects may manifest themselves over a long sequence of state-action pairs, making them hard to pinpoint. In this paper, we propose a method to link transitions with significant deviations in state with unusually large variations in subsequent rewards. Such transitions are marked as possible causal effects, and the corresponding state-action pairs are added to a separate replay buffer. In addition, we include \textit{contrastive} samples corresponding to transitions from a similar state but with differing actions. Including this Contrastive Experience Replay (CER) during training is shown to outperform standard value-based methods on 2D navigation tasks. We believe that CER can be useful for a broad class of learning tasks, including for any off-policy reinforcement learning algorithm.
\end{abstract}



\section{Introduction}

Recent work on reinforcement learning (RL) algorithms has recognised the flaws in using observed associations rather than causal reasoning for training \cite{gershman2017reinforcement, madumal2020explainable}. Several variants of causal influences in RL problems have been considered, such as the impact of an action conditioned on the current state \cite{seitzer2021causal}, the use of external observations to learn causal models \cite{gasse2021causal}, and specific observation of interventions to compute magnitude of influence \cite{dasgupta2019causal}. One may also use prior work on identification of a latent influencer \cite{shankar2018generalizing} to predict the behaviour of the environment in a given episode. In this paper, we consider a different setting (explained below) where the effect of influencers is trajectory dependent.

\begin{figure}
    \centering
    \includegraphics[width=0.4\textwidth]{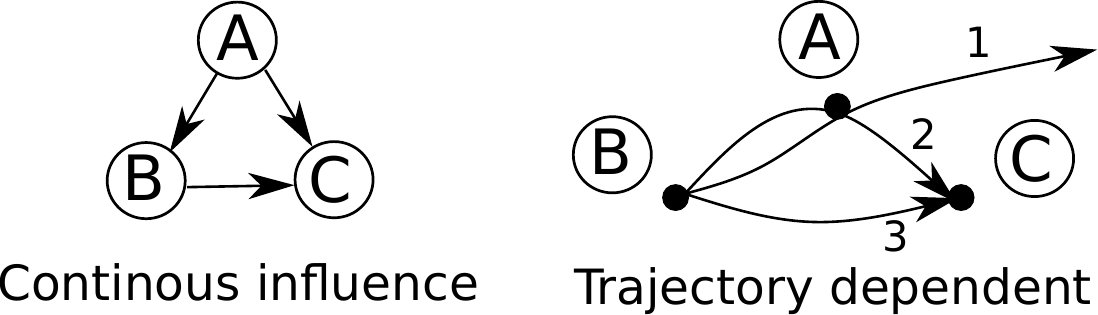}
    \caption{Difference between a one-step inference problem and a sequential decision-making problem. In the latter case, the influence of A on the long term rewards is more subtle.}
    \label{fig:causal}
\end{figure}

\textbf{Motivation: }In an exemplary causal reasoning problem (Figure \ref{fig:causal} left), the confounding influence of \texttt{A} on the relationship \texttt{B}$\rightarrow$\texttt{C} is manifested in every measurement. By contrast, the Figure \ref{fig:causal} (right) shows a scenario where the influence of \texttt{A} is manifested only in trajectories that visit \texttt{B}$\rightarrow$\texttt{A}$\rightarrow$\texttt{C} (for example, trajectory 2 in the figure). It is not apparent in trajectory 1 which visits \texttt{A} but then does not visit \texttt{C}, nor in trajectory 3 which goes \texttt{B}$\rightarrow$\texttt{C} but avoids \texttt{A}. 
{
\begin{itemize}
    \item In such sequential problems where transitions and consequent rewards are arbitrarily far apart in the temporal sense, the standard notions of causality are difficult to apply.
    \item Furthermore, rewards may be predicated on the presence or absence of a specific type of transition (such as visiting a subgoal), and thus may not manifest in every experiment.
    \item Finally, the agent may not have access to a formal structural causal model (SCM) of the environment, and must infer all causal relationships simply by analysing past trajectories.
\end{itemize}
}

\textbf{Related work: }The majority of prior work in the area of causality-aware reinforcement learning is focused on learning explicit influence models \cite{dasgupta2019causal, 
gasse2021causal, seitzer2021causal}. A survey of recent work  \cite{grimbly2021causal} highlights studies that focus on aspects of credit assignment in multi-agent settings \cite{gershman2017reinforcement} and on learning structural causal models from data \cite{madumal2020explainable}. However, these studies require variables of interest to be externally defined, following which the relationships between them may be established. As explained earlier, this assumption could be too restrictive in RL settings with complex influences. 

{\textbf{Our approach:}} We mark possible causal links through a purely internal analysis of the state-action sequence, with an emphasis on contrasting atypical trajectories (with unusually high or low rewards) with nominal trajectories. {We further posit that such reward deviations are a result of corresponding large deviations (as formalised in the next section) in state.} We add such transitions to a contrastive experience replay (CER), inspired by the concept of prioritised experience replay (PER) \cite{schaul2015prioritized}. Unlike PER which prioritises training based on prediction errors, the proposed CER contains samples with significantly different reward outcomes. The similarly named counterfactual experience replay \cite{menglin2020new} is nevertheless quite different from the proposed concept, because it involves computing target values based on future actions from a counterfactual policy. { Another study on Contrastive Reinforcement Learning \cite{poesia2021contrastive} also uses contrastive samples, but in a very different way. They assume that a problem is either `solved' or not, and the agent is aware of the transitions responsible for each outcome. Entire successful trajectories are tagged as positive samples. In our case, we do not assume any prior knowledge about the best possible rewards in an environment. Furthermore, it is quite possible for a single trajectory to contain productive as well as counterproductive transitions.}

Our proposal makes no change to the target values stored in memory. Similarly, while we use state similarity measures from literature on directed exploration \cite{badia2020never}, we do not directly influence exploration. We also do not modify any samples in the memory unlike Hindsight Experience Replay \cite{andrychowicz2017hindsight}. 

\textbf{Contribution: }We define a simple extension to offline RL algorithms, focusing on identifying key portions of trajectories and emphasising training on these samples. This is an intuitive extension of human thinking where one tends to be influenced by \textit{memorable} instances of history in greater proportion than regular events. In this preliminary work, we demonstrate the concept using Deep Q Networks \cite{mnih2015human} and Monte-Carlo style rewards on a 2D navigation task. However, the proposed method should work with almost any off-policy RL algorithm and application.

\section{Problem and Methodology}

\subsection{Formulation as MDP}

Consider a Markov Decision Process $(S,A,R,T,\gamma)$ where $S$ denotes states, $A$ denotes actions, $R$ denotes scalar rewards, $T:S\times A\rightarrow S$ is a (possibly stochastic) transition function, and $0\leq \gamma \leq 1$ is a discount factor on future rewards. Given a state $s_t$ at time step $t$, the goal of an RL algorithm is to train a policy $\pi:S\rightarrow A$ that maximises the return $G_t=\sum_{k=0}^{n} \gamma^k r_{t+k} + R_T$, where $R_T$ is a terminal reward and $n$ is the number of intermediate steps remaining in the episode. The challenge with causal reasoning in such problems is that while the action $a_t$ is computed at time $t$, it can affect rewards $r_{t'}$ for any $t'\geq t$. The consequent credit assignment problem \cite{sutton1984temporal} can sometimes be resolved with reward shaping \cite{grzes2009theoretical}. However, changing the reward structure offers no guarantees about maximisation of the original reward during training.

\subsection{Environment used in this study}

\begin{figure}[t]
    \centering
    \includegraphics[width=0.4\textwidth]{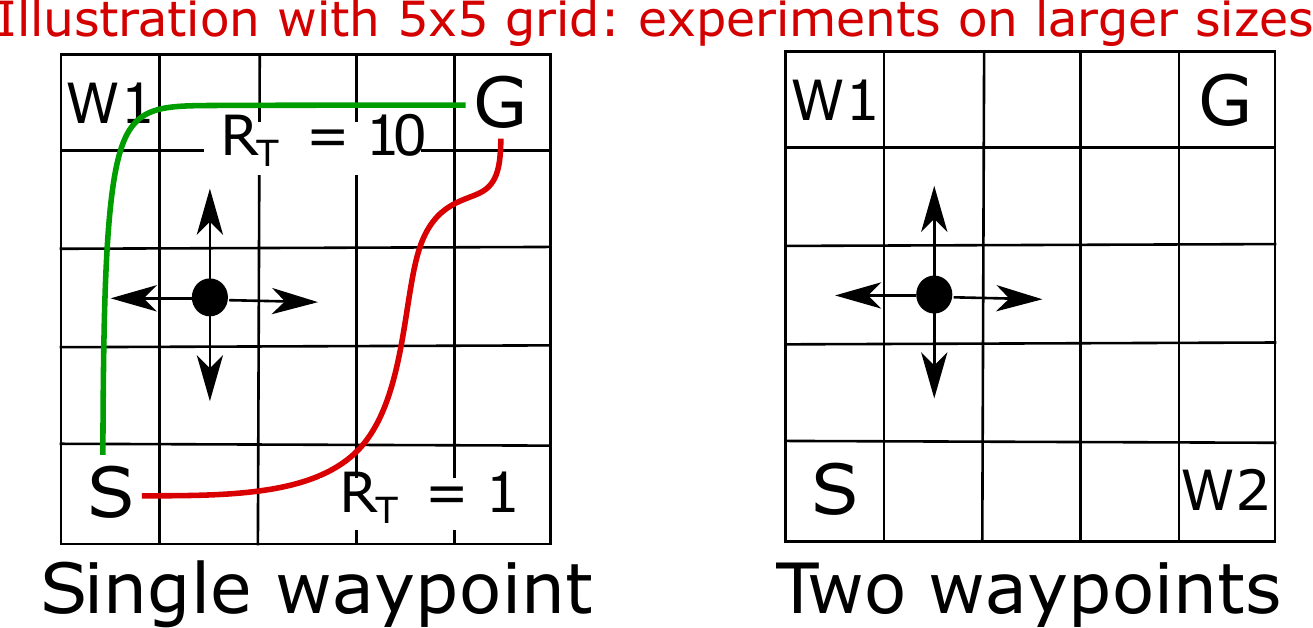}
    \caption{Schematic of the environments used in this work. Waypoints \texttt{W1} and \texttt{W2} influence the terminal rewards if they are visited on the way to \texttt{G}. Size of the environment is illustrative; experiments use 8x8 and 12x12 grids.}
    \label{fig:env}
\end{figure}

We demonstrate the effectiveness of Algorithm \ref{alg:pseudo} on a 2D grid navigation environment, as illustrated in Figure \ref{fig:env}. Consider the single waypoint version on the left. In every episode, the agent spawns at the bottom-left corner (position \texttt{S}). The actions available in each time step are \{up, down, left, right\}. A reward of $-0.01$ is provided for every step. The agent receives a terminal reward $R_T=1$ if it reaches goal \texttt{G} within $n$ time steps (depending on experiment), {similar to the standard MiniGrid environment \cite{maxime}}. However, if it visits waypoint \texttt{W1} before reaching the goal, it receives a further terminal reward of $+9$. The maximum terminal reward is thus $R_T=10$. No terminal reward is provided if the agent does not reach \texttt{G} in $n$ time steps, regardless of visiting the waypoint. In the two waypoint case, \texttt{W1} and \texttt{W2} can have different effects on the terminal reward as per the desired complexity of the learning task. These variations are described in Section \ref{sec:results}.

\begin{table}
\centering
\caption{Parameters of experiments run in this study. In all cases, a terminal reward $R_T$ of +1 is provided for going directly to the goal \texttt{G} without visiting any waypoints. Timeout $n$ and $\epsilon$-decay vary based on complexity of the learning task. All experiments are run for a fixed set of 10 random seeds.}
\label{tab:experiments}
\resizebox{.45\textwidth}{!}{
\begin{tabular}{|c|c|c|c|c|c|c|}
\hline
& Grid & $R_T$ if only & $R_T$ if only & $R_T$, both &   & Decay \\
\# & Size & visited \texttt{W1} & visited \texttt{W2} & visited & $n$ & $\epsilon$ per \\
& & before \texttt{G} & before \texttt{G} & before \texttt{G} & & episode \\
\hline
1 & 8 & 10 & - & - & 100 & 0.995 \\
2 & 8 & 2 & 2 & 10 & 160 & 0.995 \\
3 & 12 & 10 & - & - & 240 & 0.999 \\
4 & 8 & -1 & -1 & 10 & 160 & 0.999 \\
\hline
\end{tabular}
}
\end{table}

\subsection{Algorithm description}
Rather than shaping the reward, the procedure outlined in Algorithm \ref{alg:pseudo} relies on passive observation of completed trajectories. The procedure is standard DQN with an experience replay, except for the lines between the `Modification' comments. {Intuitively, we
\begin{itemize}
    \item find the largest state deviations within individual transitions in an episode, using a distance function $\Delta(s_t,s_{t+1})$,
    \item check whether the reward target for these transitions is unusually high or low (top or bottom $\psi$ percentile),
    \item if so, add these transitions to the CER, along with contrastive samples that have the same initial state $s_t$ but a different action $a \neq a_t$, and
    \item during training, generate batches by sampling both from CER and the normal buffer.
\end{itemize}
}

\begin{algorithm}[t]
 \KwData{Markov Decision Process $(S,A,R,T,\gamma)$}
 \KwResult{Trained policy $\pi:S\rightarrow A$}
 Hyperparameters: thresholds $\delta$ and $\psi$, batch size $\kappa$\;
 Initialise: $mem\_buf\leftarrow \Phi$, $CER\_buf\leftarrow \Phi$\;
 \For{$ep$ in [1:episodes]}{
  $t\leftarrow 0$, $r_0\leftarrow 0$, $ep\_buf\leftarrow \Phi$, $done \leftarrow False$\;
  \While{not done}{
  	get state $s_t$ and reward $r_{t}$ from environment\;
  	\If{$t>0$}{
  		append $(s_{t-1},a_{t-1},r_t)$ to $ep\_buf$\;
  	}
  	implement action $a_t$ from $\pi$ in $\epsilon$-greedy fashion\;
  	\If{terminal condition met}{
  		compute targets $\tau_t$ for all samples in $ep\_buf$\;
  		extend $mem\_buf$ with samples in $ep\_buf$\;
  		$done\leftarrow True$\;
  	}
  	$t\leftarrow t+1$\;
  }
  // \textbf{Modification to standard procedure starts}\;
  compute distance $\Delta(s_t,s_{t+1})\;\forall$ samples in $ep\_buf$\;
  \For{$t$ s.t. $\Delta(s_t,s_{t+1})\geq \delta\max_{k\in episode}\Delta(s_k,s_{k+1})$}{
  	\If{$\tau_t\in$ top or bottom $\psi$ percentile of mem\_buf}{
  		append $(s_t,a_t,\tau_t)$ to $CER\_buf$\;
  		find sample with $s\approx s_t$ and $a\neq a_t$ in $mem\_buf$\;
  		add this contrastive sample to $CER\_buf$\;
  	}
  }
  {get $\zeta=(0.25-0.15\cdot \epsilon)\kappa$ samples from $CER\_buf$}\;
  append $(\kappa-\zeta)$ samples from $mem\_buf$\;
  // \textbf{Modification to standard procedure ends}\;
  train on the batch of $\kappa$ samples and update $\pi$\;
 }
 \caption{Training with CER in the illustrative case of Monte-Carlo style rewards and discrete actions. The location of training calls is indicative, and need not be only at the end of an episode. We use $\delta=0.99$, $\psi=10$, $\kappa=2^{12}=4096$ for all experiments here, {and  the proportion of CER samples increases from 10\% to 25\% during training as $\epsilon$ decays from 1 to 0.}}
 \label{alg:pseudo}
\end{algorithm}


\textbf{Specifications: }The states in the single-waypoint case are given by $s_t=(\bar{x}_t,\bar{y}_t,v1)$ where $(\bar{x}_t,\bar{y}_t)$ is the normalised position on the grid, while $v1$ is a binary flag indicating whether \texttt{W1} has been visited in the current episode. This is a necessary element to ensure the Markov nature of the problem. We choose the distance function $\Delta$ to be the 2-norm $|s_{t+1}-s_t|_2$, the threshold $\delta=0.99$, discount factor $\gamma=0.99$. All algorithms use a single online network with 3 inputs, hidden layers of size (32, 8) and 4 outputs. Hidden neurons have \texttt{ReLU} activation while output neurons are \texttt{linear}. The optimiser is \texttt{Adam} with \texttt{MSE} loss and a learning rate of 0.01. Maximum buffer sizes are 2e16, with a batch size during training of 2e12. Target values are computed using Monte-Carlo backup.

{
\textbf{Analysing the distance metric $\Delta$: }Depending on the environment, one could use any relevant distance metric in the state space or the embedding space. One may also consider evaluating deviation $\Delta$ over longer sub-trajectories. The definition of a suitable $\Delta$ can be discovered in the following ways: (i) changes in context variables associated with an environment, (ii) artificial one-hot demarcation of portions of the state space, or (iii) comparison of learnt latent representations of the state instead of the raw states.

\textbf{How it works: }We can be reasonably sure that not every sample in the typical DQN buffer contains equally useful information. Given the low probability of finding extraordinary trajectories during random exploration, the bulk of samples will contain trajectories with similar outcomes. Drawing training batches from such a buffer will provide shallow gradients, and may even result in incomplete training if the exploration decays quickly. Using the CER focuses on samples with `high information', the ones which likely \textit{caused} a major deviation in reward. A significant proportion of such samples during each training run result in higher gradients at the most interesting portions of the state space.}

\textbf{Generalisation: }While Algorithm \ref{alg:pseudo} describes Monte-Carlo style rewards, one can redefine the samples appropriately for temporal difference methods (although the backup of deviated targets could take longer). One can also generalise to continuous action spaces by considering a (dis)similarity measure for actions. 

\section{results} \label{sec:results}

\textbf{Baselines:} The first baseline is vanilla DQN \cite{mnih2015human}. The second baseline is Prioritized Experience Replay \cite{schaul2015prioritized} or PER, which draws training samples in proportion to their prediction errors. 
Importance sampling is used with a correction hyper-parameter $\beta$ \cite{schaul2015prioritized}, set to increase from $0.4$ to $1.0$ linearly over the number of training episodes (after tuning). We have used the widely accepted value of $\alpha$ (prioritization) as $0.6$. The third baseline, which we call CER-nocontrast, is identical to CER except it only adds positive samples to the CER buffer and omits the contrastive samples. 

\begin{figure}
    \centering
    \includegraphics[width=0.46\textwidth]{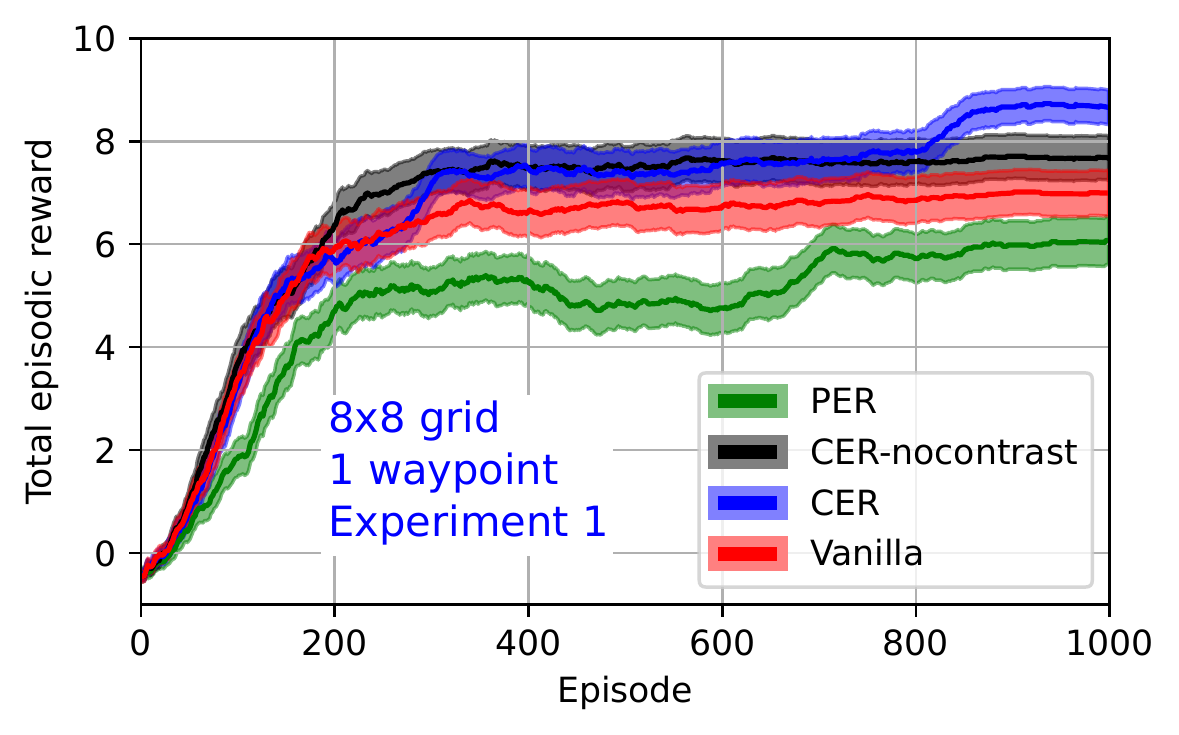}
%
    \includegraphics[width=0.46\textwidth]{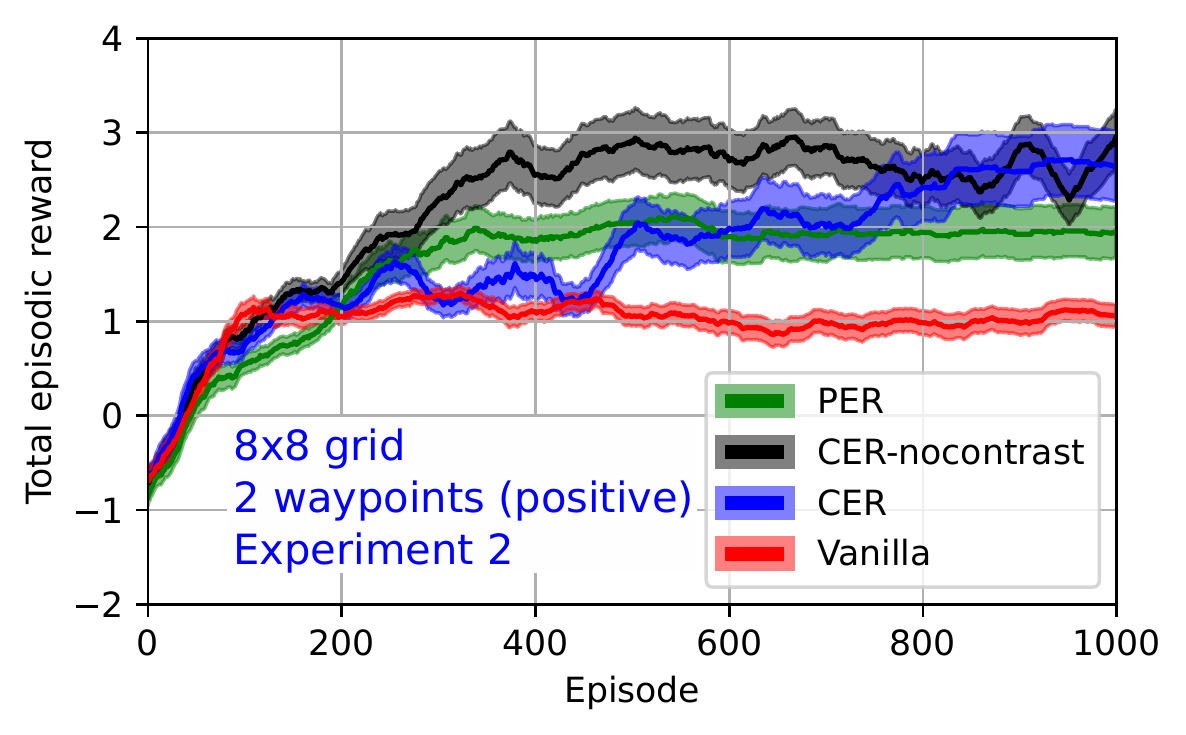}
%
    \includegraphics[width=0.46\textwidth]{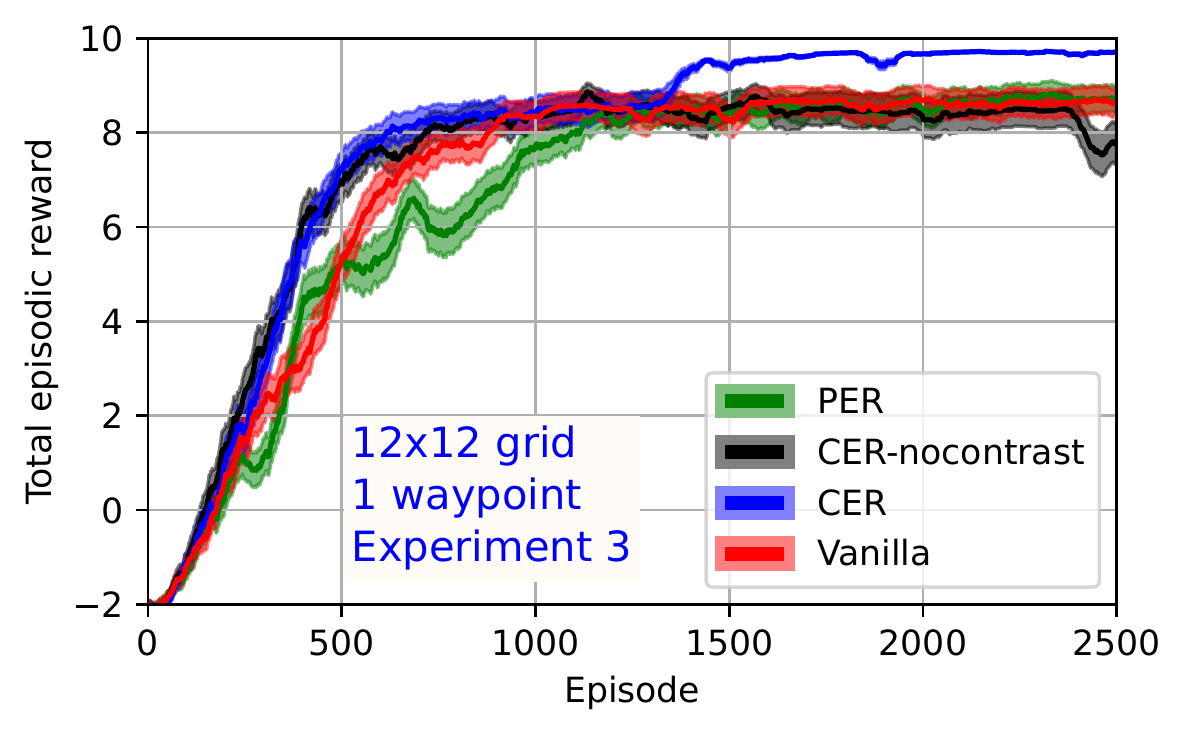}
%
    \includegraphics[width=0.46\textwidth]{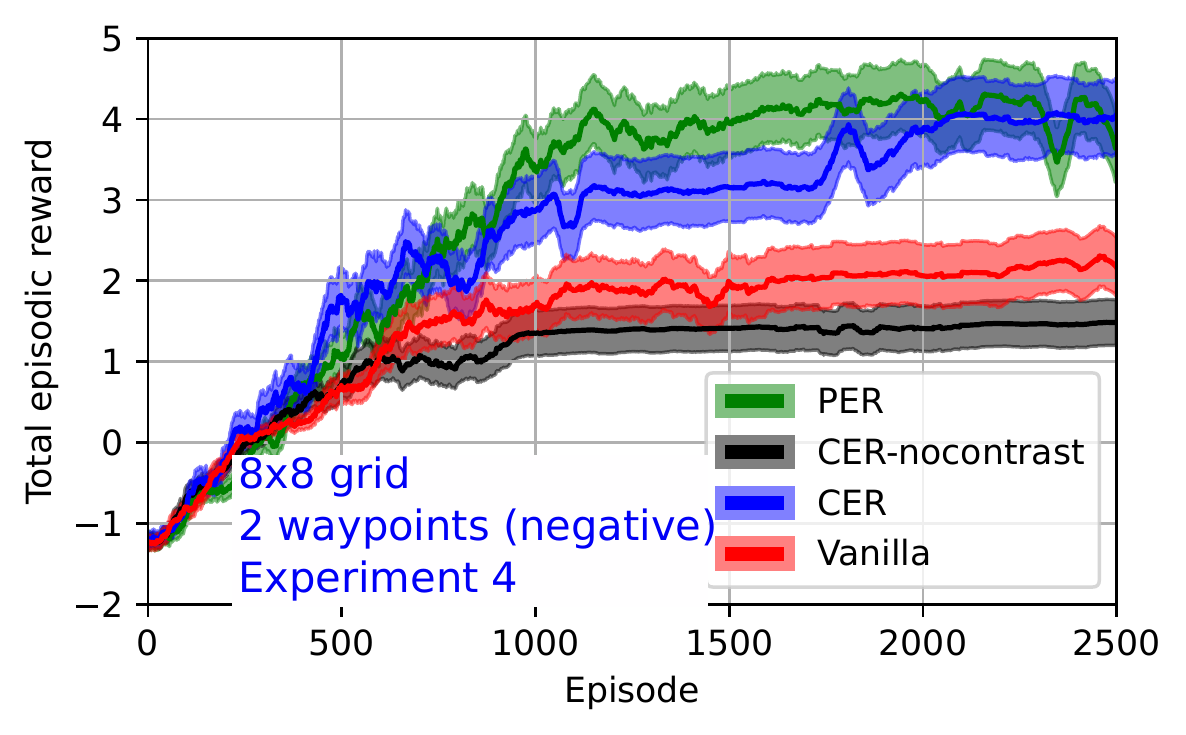}
    \caption{Training results for experiments listed in Table \ref{tab:experiments}. CER (in blue) is the method proposed in this work. Shaded regions indicate 10\% of standard deviation across 10 random seeds and a moving window of $\pm$25 episodes.}
    \label{fig:comparison}
\end{figure}

To ensure fair comparisons, all four algorithms in all experiments are run on the same set of 10 random seeds, with the same memory buffer and batch sizes, and for the same number of episodes. Figure \ref{fig:comparison} shows the results, with experimental parameters listed in Table \ref{tab:experiments}. 

\textbf{Experiment 1 [Basic single-waypoint navigation]: }We first check the efficiency of all four algorithms in a relatively simple task, that of finding a route via \texttt{W1} to the goal \texttt{G} in an {\color{red}8x8 grid}. As listed in Table \ref{tab:experiments}, the timeout $n$ is set to 100 steps and an $\epsilon$-greedy exploration strategy is used with $\epsilon$ multiplied by 0.995 after every episode. 
The first plot in Figure \ref{fig:comparison} shows that CER outperforms vanilla DQN as well as PER, and is also clearly better than CER without contrastive samples. Note that the best possible reward in this environment is 9.86.

\textbf{Experiment 2 [Two waypoints with positive effects]: }In the second {\color{red}8x8} test, we use the two-waypoint environment. The terminal reward is +1 for going to \texttt{G} directly, +2 if only one of \texttt{W1} or \texttt{W2} is visited on the way, and +10 if both waypoints are visited before \texttt{G}. We also make a small extension to the state, defining it as $s_t=(\bar{x}_t,\bar{y}_t,v1,v2)$, with $v2$ indicating whether the agent has visited \texttt{W2} in the current episode. Despite a longer timeout of $n=160$, we see the final reward being much lower than the optimal value of 9.72. The two versions of CER again outperform baselines.

\textbf{Experiment 3 [Single-waypoint on large grid]: }The third experiment involves the same rules as Experiment 1, but on a larger {\color{red}12x12 grid}. We found that the convergence was extremely poor with $n=100$ and $\epsilon$-decay of 0.995. Therefore, we increased both values as listed in Table \ref{tab:experiments}. The number of training episodes is also higher. With these changes, every random seed for CER converges to the optimal value of 9.78, while the other 3 algorithms are lower.

\textbf{Experiment 4 [Two waypoints with negative effects]: }This is also a hard exploration task, with an {\color{red}8x8 grid} and two subgoals. Unlike Experiment 2, we provide a negative terminal reward of -1 for visiting just one of \texttt{W1} or \texttt{W2} on the way to \texttt{G}, a terminal reward of +1 for going to \texttt{G} directly, and +10 for visiting both waypoints before going to the goal. CER and PER converge to similar values. This experiment also emphasises the importance of contrastive samples, given that visiting \texttt{W1} could lead either to very high or very low rewards. CER-nocontrast has access to very good and very bad samples, but no neutral ones. It converges below vanilla DQN.

\begin{figure}
    \centering
    \includegraphics[width=0.47\linewidth]{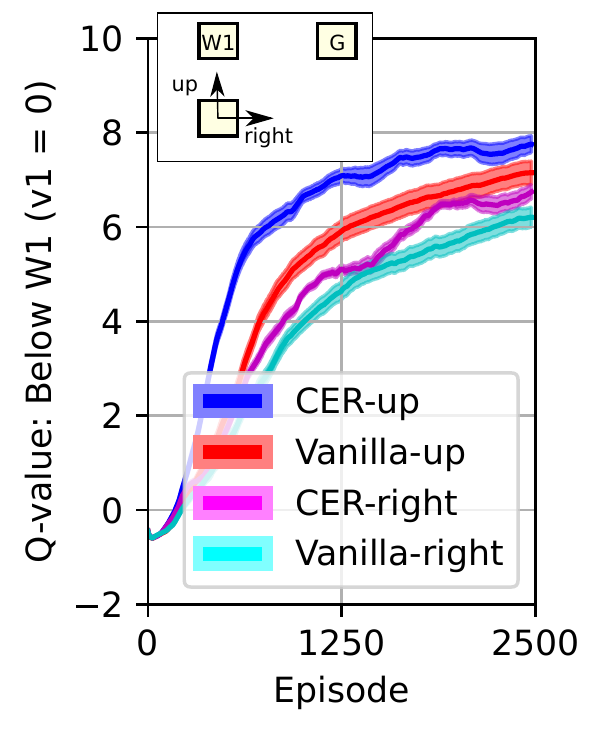}
    \includegraphics[width=0.47\linewidth]{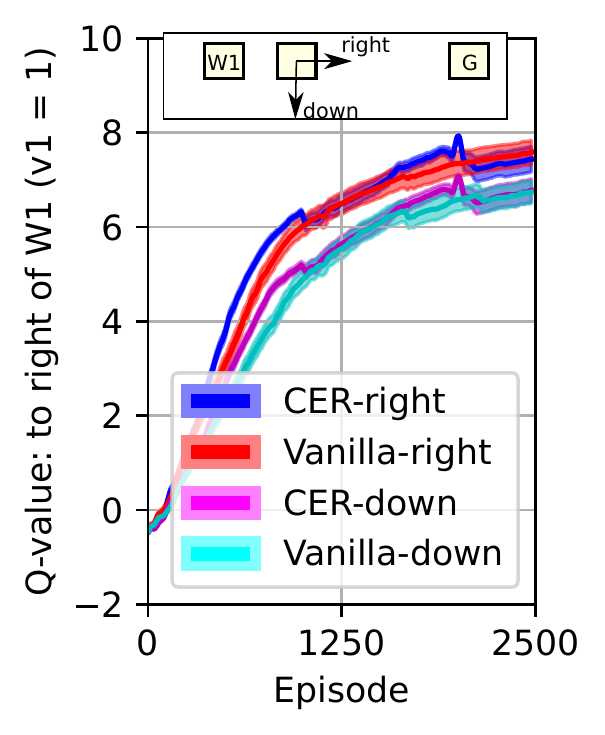}
    \caption{{Convergence of q-values in two situations for the 12x12 grid. In the first plot, action values for \texttt{UP} and \texttt{RIGHT} from the cell just below \texttt{W1} with $v1=0$. Note the clear difference between CER and vanilla algorithms. In the second plot, action values for \texttt{RIGHT} and \texttt{DOWN} from the cell just to the right of \texttt{W1}, having visited \texttt{W1} already ($v1=1$). In this case, the convergence is at the same rate for both algorithms.}}
    \label{fig:test_causal}
\end{figure}

{
\textbf{Experiment 5 [Effect of CER on key transition q-values]: }Figure \ref{fig:test_causal} plots the convergence of action values in the 12x12 sized grid (same as Experiment 3), for two specific positions. In the first plot, we consider the cell just below \texttt{W1} and two actions, \texttt{UP} and \texttt{RIGHT}. The former action leads the agent to the waypoint, while the latter avoids the waypoint. Both transitions are present in the CER buffer, and we note the clear discrimination in values as a result. The convergence is also much faster than vanilla DQN. No such distinction is observed in the second plot, where the agent has already visited \texttt{W1} and is now one cell to the right. In this case, the rates of convergence and the distinction between actions are similar for both algorithms.
}

\section{Discussion and Conclusion} \label{sec:discussion}

The key \textbf{advantage} of the Contrastive Experience Replay (CER) is the simplicity of implementation. Our proposal does not shape or re-label the rewards in any way, and only emphasises training on selected samples. The intuition aligns with human perception of past events, where significant incidents tend to stand out in memory and are frequently associated with cause-effect relationships. 

Focusing on reinforcement learning settings, we believe that CER is \textbf{generalisable} to almost all offline RL algorithms. %
Finally, since CER does not change the underlying MDP in any way, its use is mathematically correct. Even an ineffective $\Delta$ should at least be as good as the vanilla version of the RL algorithm. However, this is only a hypothesis which needs to be tested in future work.


\bibliographystyle{ACM-Reference-Format}
\bibliography{acmart}

\end{document}